%% file: main.tex
\documentclass[11pt,a4paper]{article}
\usepackage{times,latexsym}
\usepackage{url}
\usepackage[T1]{fontenc}
\usepackage{amsthm}
\usepackage{breqn}

    \usepackage[acceptedWithA]{tacl2018v2}
\theoremstyle{definition}
\newtheorem{definition}{Definition}[section]

\usepackage{xspace,mfirstuc,tabulary}
\usepackage{booktabs}
\usepackage{graphicx}
\usepackage{amssymb}
\usepackage{soul}
\usepackage[normalem]{ulem}
\usepackage{multirow}
\usepackage{amsmath}
\usepackage{array}
\usepackage{enumitem}
\usepackage{arydshln}

\usepackage[ruled,vlined]{algorithm2e}

\usepackage[normalem]{ulem}
\useunder{\uline}{\ul}{}

\usepackage{subfiles} 

\newif\iftaclinstructions

\taclinstructionsfalse 
\iftaclinstructions
\renewcommand{\confidential}{}
\renewcommand{\anonsubtext}{(No author info supplied here, for consistency with
TACL-submission anonymization requirements)}
\newcommand{\instr}
\fi

\iftaclpubformat 

\else

\fi

\title{Diff-Explainer: Differentiable Convex Optimization for\\  Explainable Multi-hop Inference}

\author{Mokanarangan Thayaparan$^{\dagger}$$^{\ddagger}$, Marco Valentino$^{\dagger}$$^{\ddagger}$, Deborah Ferreira$^{\dagger}$, \\ \textbf{Julia Rozanova}$^{\dagger}$, \textbf{Andr\'e Freitas}$^{\dagger}$$^{\ddagger}$ \\  Department of Computer Science, University of Manchester, United Kingdom$^{\dagger}$ \\  Idiap Research Institute, Switzerland$^{\ddagger}$ \\ {\tt \{firstname.lastname\}@manchester.ac.uk} \\}

\date{}

\begin{document}
\maketitle
\begin{abstract}

This paper presents \textit{Diff-}Explainer, the first \textit{hybrid} framework for explainable \textit{multi-hop inference} that integrates explicit constraints with neural architectures through differentiable convex optimization. Specifically, \textit{Diff-}Explainer allows for the fine-tuning of neural representations within a constrained optimization framework to answer and explain multi-hop questions in natural language. To demonstrate the efficacy of the hybrid framework, we combine existing ILP-based solvers for multi-hop Question Answering (QA) with Transformer-based representations. An extensive empirical evaluation on scientific and commonsense QA tasks demonstrates that the integration of explicit constraints in a end-to-end differentiable framework can significantly improve the performance of non-differentiable ILP solvers (8.91\% - 13.3\%). Moreover, additional analysis reveals that \textit{Diff-}Explainer is able to achieve strong performance when compared to standalone Transformers and previous multi-hop approaches while still providing structured explanations in support of its predictions. 
\end{abstract}

\subfile{sections/introduction.tex}

\subfile{sections/related.tex}
\subfile{sections/approach.tex}
\subfile{sections/evaluation.tex}

\subfile{sections/conclusion.tex}

\bibliography{main}
\bibliographystyle{acl_natbib}

\newpage
\subfile{sections/appendices.tex}

\end{document}

%% file: sections/introduction.tex
\section{Introduction}

\begin{figure}
    \centering
    \includegraphics[width=0.9\linewidth]{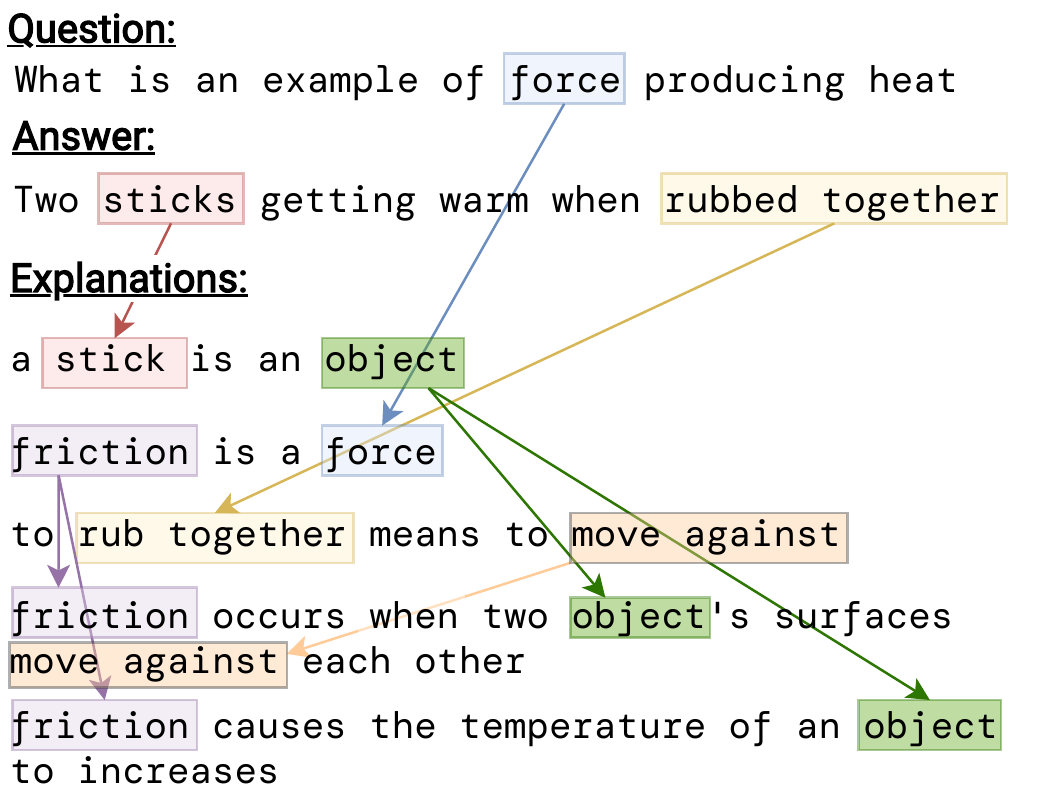}
    \caption{Example of a multi-hop QA problem with an explanation represented as a graph of multiple interconnected sentences supporting the answer \cite{xie2020worldtree,jansen2018worldtree}.}
    \label{fig:example}
\end{figure}

Explainable Question Answering (QA) in complex domains is often modelled as a \textit{multi-hop inference} problem~\cite{thayaparan2020survey,valentino2021hybrid,jansen2021textgraphs}. In this context, the goal is to answer a given question through the construction of an explanation, typically represented as a graph of multiple interconnected sentences supporting the answer (Figure~\ref{fig:example}).  ~\cite{khashabi2018question,jansen2018multi,kundu-etal-2019-exploiting,thayaparan2021explainable}.


However, explainable QA models exhibit lower performance when compared to state-of-the-art approaches, which are generally  represented by Transformer-based architectures \cite{khashabi2020unifiedqa,devlin2019bert,khot2020qasc}. While Transformers are able to achieve high accuracy due to their ability to transfer linguistic and semantic information to downstream tasks, they are typically regarded as black-boxes~\cite{liang2021explaining}, posing concerns about the interpretability and transparency of their predictions~\cite{rudin2019stop,guidotti2018survey}.

To alleviate the aforementioned limitations and find a better trade-off between explainability and inference performance,
this paper proposes \textit{Diff-}Explainer ($\partial$-Explainer), a novel \textit{hybrid} framework for multi-hop and explainable QA that combines constraint satisfaction layers with pre-trained neural
representations, enabling end-to-end differentiability.  

Recent works have shown that certain convex optimization problems can be represented as individual
layers in larger end-to-end differentiable networks~\cite{agrawal2019differentiable,diffcp2019,amos2017optnet}, demonstrating that these layers can be adapted to encode constraints and dependencies between hidden states that are hard to capture via standard neural networks. 

In this paper, we build upon this line of research, showing that convex optimization layers can be integrated with Transformers to improve explainability and robustness in multi-hop inference problems. To illustrate the impact of end-to-end differentiability, we integrate the constraints of existing ILP solvers (i.e., TupleILP~\cite{khot2017answering}, ExplanationLP~\cite{thayaparan2021explainable}) into a hybrid framework. 
Specifically, we propose a methodology to transform existing constraints into differentiable convex optimization layers and
subsequently integrate them with pre-trained sentence embeddings based on Transformers~\cite{reimers2019sentence}.

To evaluate the proposed framework, we perform extensive experiments on complex multiple-choice QA tasks requiring scientific and commonsense reasoning~\cite{clark2018think,xie2020worldtree}. In summary, the contributions of the paper are as follows:


\begin{enumerate}[noitemsep]
    \item  A novel differentiable framework for multi-hop inference that incorporates constraints via convex optimization layers into broader Transformer-based architectures. 
    \item An extensive empirical evaluation demonstrating that the proposed framework allows end-to-end differentiability on downstream QA tasks for both explanation and answer selection, leading to a substantial improvement when compared to non-differentiable constraint-based and transformer-based approaches. 
    \item We demonstrate that \textit{Diff-}Explainer is more robust to distracting information in addressing multi-hop inference when compared to Transformer-based models.
\end{enumerate}


%% file: sections/related.tex
\section{Related Work}
\label{sec:related_work}

\paragraph{Constraint-Based Multi-hop QA Solvers:} ILP has been employed to model structural and semantic constraints to perform multi-hop QA. TableILP~\cite{khashabi2016question} is one of the earliest approaches to formulate the construction of explanations as an optimal sub-graph selection problem over a set of structured tables and evaluated on multiple-choice elementary science question answering. In contrast to TableILP, TupleILP~\cite{khot2017answering} was able to perform inference over free-form text by building semi-structured representations using Open Information Extraction. SemanticILP~\cite{khashabi2018question} also comes from the same family of solvers that leveraged different semantic abstractions, including semantic role labelling, named entity recognition and lexical chunkers for inference. In contrast to previous approaches,~\citet{thayaparan2021explainable} proposed the ExplanationLP model that is optimized towards answer selection via Bayesian optimization. ExplanationLP was limited to fine-tuning only nine parameters as it is intractable to finetune large models using Bayesian optimization.

\paragraph{Hybrid Reasoning with Transformers} A growing line of research focuses on adopting Transformers for interpretable reasoning over text~\cite{clark2021transformers,gontier2020measuring,saha2020prover,tafjord2021proofwriter}. For example, \citet{saha2020prover} introduced the PROVER model that provides an interpretable transformer-based model that jointly answers binary questions over rules while generating the corresponding proofs. These models are related to the proposed framework for exploring hybrid architectures. However, these models assume that the rules are fully available in the context and are still mostly applied on synthetically generated datasets. In this paper, we take a step forward in this direction proposing an hybrid model for addressing scientific and commonsense QA which require the construction of complex explanations through multi-hop inference on external knowledge bases.


\paragraph{Differentiable Convex Optimization Layers:} Our work is in line with previous works that have attempted to incorporate optimization as a neural network layer. These works have introduced differentiable modules for quadratic problems~\cite{donti2017task,amos2017optnet}, satisfiability solvers~\cite{wang2019satnet} and submodular optimizations~\cite{djolonga2017differentiable,tschiatschek2018differentiable}. Recent works also offer differentiation through convex cone programs~\cite{busseti2019solution,diffcp2019}. In this work, we use the differentiable convex optimization layers proposed by~\citet{agrawal2019differentiable}. These layers provide a way to abstract away from the conic form, letting users define convex optimization in natural syntax. The defined convex optimization problem is converted by the layers into a conic form to be solved by a conic solver~\cite{odonoghue:21}.

%% file: sections/approach.tex
\section{Multi-hop Question Answering via Differentiable Convex Optimization}
\label{problem_formulation}

\begin{figure*}[t]
    \centering
    \resizebox{\textwidth}{!}{\includegraphics{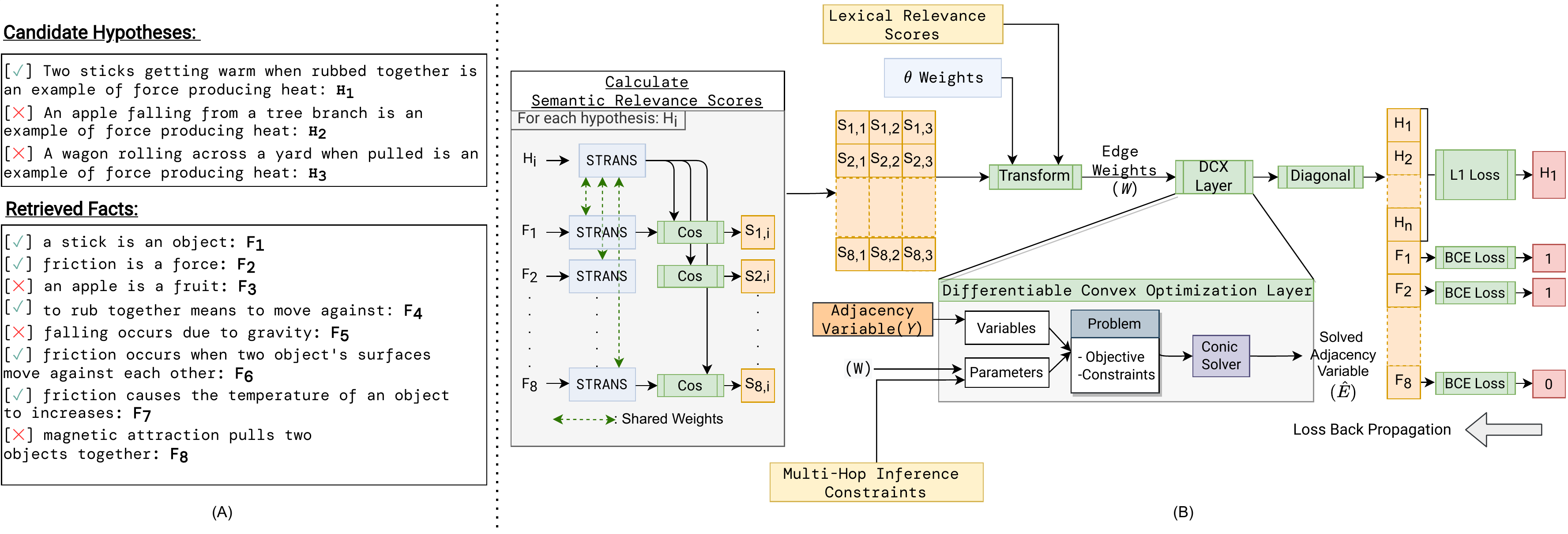}}
    \caption{ Overview of our approach: Illustrates the end-to-end architectural diagram of \textit{Diff-}Explainer for the provided example.}
    \label{fig:end_to_end}
\end{figure*}


The problem of Explainable Multi-Hop Question Answering can be stated as follows: 

\begin{definition}[\textit{Explanations in Multi-Hop Question Answering}] Given a question $Q$, answer $a$ and a knowledge base $F_{kb}$  (composed of natural language sentences), we say that we may \emph{infer}  hypothesis $h$ (where   hypotheses $h$ is the concatenation of $Q$ with $a$) if there exists a subset ($F_{exp}$) of supporting facts $\{f_1, f_2, \ldots \} \subseteq F_{kb}$ of statements which would allow a human being to deduce $h$ from $\{f_1, f_2, \ldots\}$. We call this set of facts an \emph{explanation} for $h$.
\end{definition}

Given a question ($Q$) and a set of candidate answers $C=\{c_1,~c_2,~c_3, ...,~c_n\}$ ILP-based approaches~\cite{khashabi2016question,khot2017answering,thayaparan2021explainable} convert them into a list of hypothesis $H = \{h_{1},~h_{2},~h_{3},~\dots,~h_{n}\}$ by concatenating question and candidate answer. For each hypothesis $h_i$ these approaches typically adopt a retrieval model (e.g: BM25, FAISS~\cite{DBLP:journals/corr/JohnsonDJ17}),  to select a list of candidate explanatory facts $F = \{f_{1},~f_{2},~f_{3},~\dots,~f_{k}\}$, and construct a weighted graph $G = (V,E,W)$ with edge weights $W: E~\rightarrow~\mathbb{R}$ where $V=~\{\{h_i\}~\cup~F\}$, edge weight $W_{ik}$ of each edge $E_{ik}$ denote how relevant a fact $f_k$ is with respect to the hypothesis $h_i$.

Based on these definitions, ILP-based QA can be defined as follows:


\begin{definition}[\textit{ILP-Based Multi-Hop QA}]
Find a subset $V^* \subseteq V$, $h \in V^*$, $V^* \setminus\{h\}=F_{exp}$  and $E^* \subseteq E$  such that the induced subgraph $G^* = (V^*,E^*)$ is connected, weight $W[G^* = (V^*,~E^*)] := \sum_{e \in E^*} W(e)$ is maximal and adheres to set of constraints $M_c$ designed to emulate multi-hop inference. The hypothesis $h_i$ with the highest subgraph weight $W[G^* = (V^*,~E^*)]$ is selected to be the correct answer $c_{ans}$.
\end{definition}

The ILP-based inference has two main challenges in producing convincing explanations. First, design edge weights $W$, ideally capturing a quantification of the relevance of the fact to the hypothesis. Second, define constraints that emulate the multi-hop inference process.




\subsection{Limitations with Existing ILP formulations}
\label{sec:limitations}

In previous work, the construction of the graph $G$ requires predetermined edge-weights based on lexical overlaps~\cite{khot2017answering} or semantic similarity using sentence embeddings~\cite{thayaparan2021explainable}, on top of which combinatorial optimization strategies are performed separately. From those approaches, ExplanationLP proposed by \citet{thayaparan2021explainable} is the only approach that modifies the graph weight function by optimizing the weight parameters $\theta$ by fine-tuning them for inference via Bayesian Optimization over pre-trained embeddings. 

In contrast, we posit that learning the graph weights dynamically by fine-tuning the underlying neural embeddings towards answer and explanation selection will lead to more accurate and robust performance.
To this end, the constraint optimization strategy should be differentiable and efficient. However, Integer Linear Programming based approaches present two critical shortcomings that prevent achieving this goal:

\begin{enumerate}[noitemsep,leftmargin=*,topsep=0pt]
    \item The Integer Linear Programming formulation operates with discrete inputs/outputs resulting in \textit{non-differentiability}~\cite{DBLP:journals/corr/abs-2105-02343}. Consequently, it cannot be integrated with deep neural networks and trained end-to-end. Making ILP differentiable requires non-trivial assumptions and approximations~\cite{DBLP:journals/corr/abs-2105-02343}.
    \item Integer Programming is known to be NP-complete, with the special case of 0-1 integer linear programming being one of Karp's 21 NP-complete problems \cite{karp1972reducibility}. Therefore, as the size of the combinatorial optimization problem increases, finding exact solutions becomes computationally intractable. This intractability is a strong limitation for multi-hop QA in general since these systems typically operate on large knowledge bases and corpora. 
\end{enumerate}


\subsection{Subgraph Selection via Semi-Definite Programming}
\label{sec:subgraph_seletion_diff}

Differentiable convex optimization (DCX) layers~\cite{agrawal2019differentiable} provide a way to encode constraints as part of a deep neural network. However, an ILP formulation is non-convex~\cite{wolsey2020integer,schrijver1998theory} and cannot be incorporated into a differentiable convex optimization layer. The challenge is to approximate ILP with convex optimization constraints. 

In order to alleviate this problem, we turn to \emph{Semi-Definite programming} (SDP)~\cite{vandenberghe1996semidefinite}. SDP is non-linear but convex and has shown to efficiently approximate combinatorial problems. 

A semi-definite optimization is a convex optimization of the form:

\begin{align}
    & minimize  & C \cdot X  \\
    &s.t & A \cdot X = b_i,~i=1,2,~\cdots,~m,\\
    &&X \succeq 0,
\end{align}

Here $X \in \mathbb{S}^{n}$ is the optimization variable and $C, A_1, \ldots, A_p \in \mathbb{S}^{n}$, and $b_1, \ldots, b_p \in \mathbb{R}$. $X \succeq 0$ is a matrix inequality with $\mathbb{S}^{n}$ denotes a set of $n \times n$ symmetric matrices.



SDP is often used as a convex approximation of traditional NP-hard combinatorial graph optimization problems, such as the max-cut problem, the dense k-subgraph problem and the quadratic $\{0-1\}$ programming problem~\cite{lovasz}.
 
Specifically, we adopt the semi-definite relaxation of the following quadratic $\{0,~1\}$ problem:

\begin{align}
    & maximize &  y^TWy \\
    & & y \in \{0, 1\}^n
\end{align}

Here $W$ is the edge weight matrix of the graph $G$ and the optimal solution for this problem $\hat{y}$ indicates if a node is part of the induced subgraph $G^*$.

We follow \citet{helmberg2000semidefinite} in their reformulation and relaxation of this problem. Instead of vectors $y \in \{ 0,1\}^n$, we optimize over the set of \emph{positive semidefinite matrices} satisfying the SDP constraint in the following relaxed convex optimization problem\footnote{See \cite{helmberg2000semidefinite} for the derivation from the original optimization problem.}:

\begin{align}
    & maximize &  \quad  \langle W,Y \rangle  \\
    &s.t & Y - diag(Y)diag(Y)^T \succeq 0 
    \label{eq:sdp_constraint}
\end{align}

where $\langle W,Y \rangle = \text{trace}(WY)$, $Y = yy^T$, $diag(Y)=y$.

The optimal solution for $Y$ in this problem   $\hat{E}~\in~[0,~1]$  indicates if an edge is part of the subgraph $G^*$. In addition to the semi-definite constraints, we also impose Multi-hop inference constraints $M_c$. These constraints are introduced in Section~\ref{sec:answer_explanation} and the Appendix.

This reformulation provides the tightest approximation for the optimization with the convex constraints. Since this formulation is convex, we can now integrate it with differentiable convex optimization layers.
Moreover, the semi-definite program relaxation can be solved by adopting the interior-point method~\cite{de2006aspects,vandenberghe1996semidefinite} which has been proved to run in polynomial time~\cite{karmarkar1984new}. To the best of our knowledge, we are the first to employ SDP to solve a natural language processing task.

         
         


\subsection{\textit{Diff-}Explainer: End-to-End Differentiable Architecture}

\textit{Diff-}Explainer is an end-to-end differentiable architecture that simultaneously solves the constraint optimization problem and dynamically adjusts the graph edge weights for better performance. We adopt \emph{differentiable convex optimization} for the optimal subgraph selection problem. The complete architecture and setup are described in the subsequent subsections and Figure ~\ref{fig:end_to_end}.

We transform a multi-hop question answering dataset into a multi-hop QA dataset by converting an example's question ($q$) and the set of candidate answers $C=\{c_1,~c_2,~c_3,~\dots,~c_n\}$ into hypotheses $H=\{h_{1},~h_{2},~h_{3},~\dots,~h_{n}\}$ (See Figure~\ref{fig:end_to_end}A) by using the approach proposed by~\citet{demszky2018transforming}. To build the initial graph,  for the hypotheses set  $H$ we adopt a retrieval model to select a list of candidate explanatory facts $F = \{f_{1},~f_{2},~f_{3},~\dots,~f_{k}\}$ to construct a weighted complete bipartite graph $G = (H,~F,~E,~W)$, where the weights $W_{ik}$ of each edge $E_{ik}$ denote how relevant a fact $f_k$ is with respect to the hypothesis $h_i$. Departing from traditional ILP approaches~\cite{thayaparan2021explainable,khashabi2018question,khashabi2016question}, the aim is to select the correct answer $c_{ans}$ and relevant explanations $F_{exp}$ with a single graph.

In order to demonstrate the impact of \textit{Diff-}Explainer, we reproduce the formalization introduced by previous ILP solvers. Specifically, we approximate the two following solvers:


\begin{itemize}[noitemsep, leftmargin=*]
    \item \textbf{TupleILP}~\cite{khot2017answering}: TupleILP constructs a semi-structured knowledge base using tuples extracted via Open Information Extraction (OIE)  and performs inference over them. For example, in Figure~\ref{fig:end_to_end}A, $F1$ will be decomposed into \textit{(a stick; is a; object)} and the subject (\textit{a stick}) will be connected to the hypothesis to enforce constraints and build the subgraph. 
    \item \textbf{ExplanationLP}~\cite{thayaparan2021explainable}: ExplanationLP classifies facts into abstract and grounding facts. Abstract facts are core scientific statements. Grounding facts help connect the generic terms in the abstract facts to the terms in the hypothesis. For example, in Figure~\ref{fig:end_to_end}A, $F_{1}$ is a grounding fact and helps to connect the hypothesis with the abstract fact $F_{7}$. The approach aims to emulate abstract reasoning.
\end{itemize}

Further details of these approaches can be found in the Appendix.



To demonstrate the impact of integrating a convex optimization layer into a broader end-to-end neural architecture, \textit{Diff-}Explainer employs a transformer-based sentence embedding model. Figure~\ref{fig:end_to_end}B describes the end-to-end architectural diagram of \textit{Diff-}Explainer. Specifically, we incorporate a differentiable convex optimization layer with Sentence-Transformer (STrans)~\cite{reimers2019sentence}, which has demonstrated state-of-the-art performance on semantic sentence similarity benchmarks. 

STrans is adopted to estimate the relevance between hypothesis and facts during the construction of the initial graph. We use STrans as a bi-encoder architecture to minimize the computational overload and operate on large number of sentences. The semantic relevance score from STrans is complemented with a lexical relevance score computed considering the shared terms between hypotheses and facts. We calculate semantic and lexical relevance as follows:

\noindent\textbf{Semantic Relevance ($s$)}: Given a hypothesis $h_i$ and fact $f_j$ we compute sentence vectors of $\vec{h_i}^{\,} = STrans(h_i)$ and $\vec{f_j}^{\,} = STrans(f_j)$ and calculate the semantic relevance score using cosine-similarity as follows:
    \begin{equation}
        \small
        s_{ij} = S(\vec{h_i}^{\,},~\vec{f_j}^{\,}) =  \frac{\vec{h_i}^{\,} \cdot \vec{f_j}^{\,}}{ \|\vec{h_i}^{\,} \|\|\vec{f_j}^{\,}\|}
    \end{equation}
    
\noindent\textbf{Lexical Relevance ($l$)}: The lexical relevance score of hypothesis $h_i$ and $f_j$ is given by the percentage of overlaps between unique terms (here, the function $trm$ extracts the lemmatized set of unique terms from the given text):
    \itemsep0em 
    \begin{equation}
        \small
        l_{ij} = L(h_i,~f_j) = \frac{\vert trm(h_i) \cap trm(f_j) \vert}{max(\vert trm(h_i)\vert, \vert trm(f_j) \vert)}
    \end{equation}
    
Given the above scoring function, we construct edge weights matrix ($W$) as follows:

\begin{multline}
\small
 W_{ij} = [\theta^s_1,~\theta^s_2,~\dots~,\theta^s_n]\cdot[s_{ij}^{\mathcal{D}_1},~s_{ij}^{\mathcal{D}_2},~\dots~,s_{ij}^{\mathcal{D}_n}] \\+  [\theta^l_1,~\theta^l_2,~\dots~,\theta^l_n]\cdot[l_{ij}^{\mathcal{D}_1},~l_{ij}^{\mathcal{D}_2},~\dots~,l_{ij}^{\mathcal{D}_n}]
\end{multline}

Here relevance scores are weighted by weight parameters ($\theta$) with $\theta$ clamped to $[0,1]$. $s_{ij}^{\mathcal{D}_k}$ (or $l_{ij}^{\mathcal{D}_k}$)  is $s_{ij}$ (or $l_{ij}$)  if $(i,j)$ satisfy condition $\mathcal{D}_k$ or $0$ otherwise. TupleILP has two weights each for lexical and semantic relevance. Meanwhile, ExplanationLP has nine weights based on the type of fact and relevance type. Additional details on how to calculate $W$ for each approach can be found in the Appendix.

\subsection{Answer and Explanation Selection}
\label{sec:answer_explanation}



Given edge variable $Y$  and node variable $y$ ($diag(Y) = y$) (See Section~\ref{sec:subgraph_seletion_diff}) where 1 means the edge/node is part of the subgraph and 0 otherwise, we design the the answer selection constraint is defined as follows:

\begin{equation}
    \begin{aligned}
    \sum_{i~\in~H} Y_{ii} & = 1 \\ 
    \end{aligned}
\end{equation}
Each entry in the edge diagonal represents a value between 0 and 1, indicating whether the corresponding node in the initial graph should be included in the optimal subgraph.


Explanation selection is done via the following constraint that limits the number of nodes in the subgraph to be $m$.
\begin{equation}
    \begin{aligned}
    \sum_{i~\in~V} Y_{ii} & = m + 1 \\ 
    \end{aligned}
\end{equation}

Apart from these functional constraints, ILP based methods also impose semantic and structural constraints. For instance, ExplanationLP places explicit grounding-abstract fact chain constraints to perform efficient abstractive reasoning and TupleILP enforces constraints to leverage the SPO structure to align and select facts. See the Appendix on how these constraints are designed and imposed within \textit{Diff-}Explainer.


The output from the DCX layer returns the solved edge adjacency matrix $\hat{E}$ with values between 0 and 1. We interpret the diagonal values of $\hat{E}$ be the probability of the specific node to be part of the selected subgraph. The final step is  to optimize the sum of the L1 loss $l_1$ between the selected answer and correct answer $c_{ans}$ for the answer loss $\mathcal{L}_{ans}$:

\begin{equation}
    \mathcal{L}_{ans} = l_1(diag(\hat{E})[h_1,~h_2,~\dots,~h_n],~c_{ans}) 
\end{equation}

As well as the binary cross entropy loss $l_b$ between the selected explanatory facts and true explanatory facts $F_{exp}$ for the explanatory loss $\mathcal{L}_{exp}$:
\begin{equation}
    \mathcal{L}_{exp} = l_b(diag(\hat{E})[f_1,~f_2,~\dots,~f_k],~F_{exp})
\end{equation} 

We add the losses to backpropagate to learn the $\theta$ weights and fine-tune the sentence transformers. The pseudo-code to train \textit{Diff-}Explainer end-to-end is summarized in Algorithm~\ref{algo}.

\SetKwComment{Comment}{/* }{ */}

\begin{algorithm}[h]
\small
\SetAlgoLined
\KwData{$M_c \gets \textrm{Multi-hop inference constraints}$}
\KwData{$Ans_c \gets \textrm{Answer selection constraint}$}
\KwData{$Exp_c \gets \textrm{Explanation selection constraint}$}
\KwData{$f_w \gets \textrm{Graph weight function}$}
$G \gets \textrm{fact-graph-construction}(H,~F)$\;
$l \gets L(H,~F)$\;
$\theta \gets clamp([0,~1])$\;
$epoch \gets 0$\;
\While{epoch $\leq$ max\_epochs}{
  $\vec{F} \gets STrans(F)$\;
  $\vec{H}  \gets STrans(H)$\;
  $s \gets S(\vec{H},~\vec{F})$\;
  $W \gets f_w(s,~l;~\theta)$\;
  $\hat{E} \gets DCX(W,~\{M_c,~Ans_c,~Exp_c\})$\;
  
  $\hat{V}\gets diag(\hat{E})$\;
  $\mathcal{L}_{ans} \gets l_1(\hat{V}[h_1,~h_2,~\dots,~h_n],~c_{ans})$\;
  \eIf{$F_{exp}$ is available}{
  $\mathcal{L}_{exp} \gets l_b(\hat{V}[f_1,~f_2,~\dots,~f_k],~F_{exp})$\;
  $loss = \mathcal{L}_{ans} +\mathcal{L}_{exp}$\;
  }{
    $loss = \mathcal{L}_{ans}$\;
  }
  update $\theta$, $STrans$ using AdamW optimizer by minimizing $loss$\;
  $epoch \gets epoch + 1$\;
  
}
\KwResult{Store best $\theta$ and $STrans$}
 \caption{\small Training \textit{Diff-}Explainer}
\label{algo}
\end{algorithm}



%% file: sections/evaluation.tex
\section{Empirical Evaluation}

\paragraph{Question Sets:}  We use the following multiple-choice question sets to evaluate the \textit{Diff-}Explainer. 

\noindent \textbf{(1) WorldTree Corpus}~\cite{xie2020worldtree}: The 4,400 question and explanations in the WorldTree corpus are split into three different subsets: \emph{train-set}, \emph{dev-set} and \emph{test-set}. We use the \emph{dev-set} to assess the explainability performance since the explanations for \emph{test-set} are not publicly available. 

\noindent \textbf{(2) ARC-Challenge Corpus}~\cite{clark2018think}:  ARC-Challenge is a multiple-choice question dataset which consists of question from science exams from grade 3 to grade 9. These questions have proven to be challenging to answer for other LP-based question answering and neural approaches. 

\paragraph{Experimental Setup}: We use \textit{all-mpnet-base-v2} model as the Sentence Transformer model for the sentence representation in \textit{Diff-}Explainer. The motivation to choose this model is to use a pre-trained model on natural language inference and MPNet$_{Base}$~\cite{DBLP:journals/corr/abs-2004-09297} is smaller compared to large models like BERT$_{Large}$, enabling us to encode a larger number of facts. 
Similarly, for fact retrieval representation, we use \textit{all-mpnet-base-v2} trained with gold explanations of WorldTree Corpus to achieve a Mean Average Precision of 40.11 in the dev-set. We cache all the facts from the background knowledge and retrieve the top $k$ facts using MIPS retrieval~\cite{DBLP:journals/corr/JohnsonDJ17}. We follow a similar setting proposed by~\citet{thayaparan2021explainable} for the background knowledge base by combining over 5000 abstract facts from the WorldTree table store (WTree) and over 100,000 \textit{is-a} grounding facts from ConceptNet (CNet)~\cite{DBLP:journals/corr/SpeerCH16}. Furthermore, we also set $m$=2 in line with the previous configurations from TupleILP and ExplanationLP\footnote{We fine-tune \textit{Diff-}Explainer using a learning rate of $1e\text{-}5$, $14$ epochs, with a batch size of $8$}.

\paragraph{Baselines:} In order to assess the complexity of the task and the potential benefits of the convex optimization layers presented in our approach, we show the results for different baselines. We run all models with $k=\{1,\ldots, 10, 25, 50, 75, 100\}$ to find the optimal setting for each baseline and perform a fair comparison. For each question, the baselines take as input a set of hypotheses, where each hypothesis is associated with $k$ facts, ranked according to the fact retrieval model. 


\noindent\textbf{(1) IR Solver}~\cite{clark2018think}: This approach attempts to answer the questions by computing the accumulated score from all $k$ obtained from summing up the retrieval scores. In this case, the retrieval scores are calculated using the cosine similarity of fact and hypothesis sentence vectors obtained from the STrans model trained on gold explanations.  The hypothesis associated with the highest score is selected as the one containing the correct answer.

\noindent\textbf{(2) BERT$_{Base}$ and BERT$_{Large}$}~\cite{devlin2019bert}: To use BERT for this task, we concatenate every hypothesis with $k$ retrieved facts, using the separator token \texttt{[SEP]}. We use the HuggingFace~\cite{DBLP:journals/corr/abs-1910-03771} implementation of \textit{BertForSequenceClassification}, taking the prediction with the highest probability for the positive class as the correct answer\footnote{We fine-tune both versions of BERT using a learning rate of $1e\text{-}5$, $10$ epochs, with a batch size of $16$ for \emph{Base} and $8$ for \emph{Large}}.

\noindent \textbf{(3) PathNet}~\cite{kundu-etal-2019-exploiting}: PathNet is a graph-based neural approach that constructs a single linear path composed of two facts connected via entity pairs for reasoning. It uses the constructed paths as evidence of its reasoning process. They have exhibited strong performance for multiple-choice science questions.

\noindent \textbf{(4) TupleILP} and \textbf{ExplanationLP}: Both replications of the non-differentiable solvers are implemented with the same constraints as \textit{Diff-}Explainer via SDP approximation without fine-tuning end-to-end; instead, we fine-tune the $\theta$ parameters using Bayesian optimization\footnote{We fine-tune for 50 epochs using the \href{https://ax.dev/}{Adpative Experimentation Platform}.} and frozen STrans representations. This baseline helps us to understand the impact of the end-to-end fine-tuning.

\subsection{Answer Selection}

\paragraph{WorldTree Corpus:} Table~\ref{tab:worldtree_answer} presents the answer selection performance on the WorldTree corpus in terms of accuracy, presenting the best results obtained for each model after testing for different values of $k$. We also include the results for BERT without explanation in order to evaluate the influence extra facts can have on the final score. We also present the results for two different training goals, optimizing for only the answer and optimizing jointly for answer and explanation selection.


\begin{table}[t]
\centering
\small
\begin{tabular}{lc}
 \toprule
\multicolumn{1}{c}{\textbf{Model}}                  & \textbf{Acc}         \\ \midrule
\multicolumn{2}{c}{\textbf{Baselines}}                                     \\
IR Solver                                           & 50.48                \\
        \midrule
BERT$_{Base}$ (Without Retrieval)                                        & 45.43             \\
BERT$_{Base}$                                         & 58.06                \\
BERT$_{Large}$ (Without Retrieval)                                      & 49.63            \\
BERT$_{Large}$                                        & 59.32                \\
        \midrule
TupleILP                                      & 49.81                \\ 
ExplanationLP & 62.57             \\ \midrule
PathNet  &       43.40       \\ \midrule
        \midrule
\multicolumn{2}{c}{\textbf{\textit{Diff-}Explainer}}                                 \\
\underline{TupleILP constraints} &                      \\
- Answer Selection only         &     61.13               \\
- Answer and explanation selection         &     63.11                \\
\underline{ExplanationLP constraints} &                      \\
- Answer selection only         &     69.73                \\
- Answer  and explanation selection         &     \textbf{71.48}                \\
\bottomrule
\end{tabular}%
\caption{Answer selection performance for the baselines and across different configurations of our approach on WorldTree Corpus.}
\label{tab:worldtree_answer}
\end{table}

We draw the following conclusions from the empirical results obtained on the WorldTree corpus (The performance increase here are in expressed absolute terms):

\noindent (1) \textit{Diff-}Explainer with ExplanationLP and TupleILP outperforms the respective non-differentiable solvers by 13.3\% and 8.91\%. This increase in performance indicates that \textit{Diff-}Explainer can incorporate different types of constraints and significantly improve performance compared with the non-differentiable version.

\noindent (2) It is evident from the performance obtained by a large model such as BERT$_{Large}$ (59.32\%) that we are dealing with a non-trivial task. The best \textit{Diff-}Explainer setting (with ExplanationLP) outperforms the best transformer-based models with and without explanations by 12.16\% and 21.85\%. Additionally, we can also observe that both with TupleILP and ExplanationLP, we obtain better scores over the transformer-based configurations.

\noindent (3) Fine-tuning with explanations yielded better performance than only answer selection with ExplanationLP and TupleILP, improving performance by 1.75\% and 1.98\%. The increase in performance indicates that \textit{Diff-}Explainer can learn from the distant supervision of answer selection and improve in a strong supervision setting. 

\noindent (4) Overall, we can conclude that incorporating constraints using differentiable convex optimization with transformers for multi-hop QA leads to better performance than pure constraint-based or transformer-only approaches.

\paragraph{ARC Corpus:} Table~\ref{tab:arc_answer} presents a comparison of baselines and our approach with different background knowledge bases: TupleInf, the same as used by TupleILP~\cite{khot2017answering}, and WorldTree \& ConceptNet as used by ExplanationLP~\cite{thayaparan2021explainable}. We have also reported the original scores reported by the respective approaches.





For this dataset, we use our approach with the same settings as the model applied to WorldTree, and fine-tune for only answer selection since ARC does not have gold explanations. Models employing Large Language Models (LLMs) trained across multiple question answering datasets like UnifiedQA~\cite{khashabi2020unifiedqa} and AristoBERT~\cite{xu2021dynamic} have demonstrated strong performance in ARC with an accuracy of 81.14 and 68.95 respectively.

To ensure a fair comparison, we only compare the best configuration of \textit{Diff-}Explainer with other approaches that have been trained \textit{only} on the ARC corpus and provide some form of explanations in Table~\ref{tab:arc_baselines}. Here the explainability column indicates if the model delivers an
explanation for the predicted answer. A subset of the approaches produces evidence for the answer but remains intrinsically black-box. These models have been marked as \textit{Partial}.

\noindent (1) \textit{Diff-}Explainer improves the performance of non-differentiable solvers regardless of the background knowledge and constraints. With the same background knowledge, our model improves the original TupleILP and ExplanationLP by 10.12\% and  2.74\%, respectively. 

\noindent (2) Our approach also achieves the highest performance for partially and fully explainable approaches trained \textit{only} on ARC corpus. 

\noindent (3) As illustrated in Table~\ref{tab:arc_baselines}, we outperform the next best fully explainable baseline (ExplanationLP) by 2.74\%. We also outperform the stat-of-the-art model AutoRocc~\cite{yadav-etal-2019-quick} (uses BERT$_{Large}$) that is only trained on ARC corpus by 1.71\% with 230 million fewer parameters.

\noindent (4) Overall, we achieve consistent performance improvement over different knowledge bases (TupleInf, Wordtree \& ConceptNet) and question sets (ARC, WorldTree), indicating that the robustness of the approach.

\begin{table}[t]
    \centering
    \small
    \resizebox{\linewidth}{!}{
    \begin{tabular}{p{3cm}p{2.5cm}c}
    \toprule
       \textbf{Model}  & \textbf{Background KB} & \textbf{Acc} \\
      \midrule
        TupleILP\newline~\cite{khot2017answering} & TupleInf & 23.83 \\
        ExplanationLP~\cite{thayaparan2021explainable} & WTree \& CNet & 40.21 \\
      \midrule
        TupleILP (Ours) & TupleInf & 29.12 \\
        ExplanationLP (Ours) & WTree \& CNet & 37.40 \\
        \midrule
        \midrule
        \multicolumn{3}{c}{\textbf{\textit{Diff-}Explainer}}\\
        TupleILP \newline Constraints & TupleInf & 33.95 \\
        ExplanationLP \newline Constraints& WTree \& CNet & \textbf{42.95} \\
        
    \bottomrule
    \end{tabular}
    }
    \caption{Answer Selection performance on ARC corpus with \textit{Diff-}Explainer fine-tuned on answer selection.}
    \label{tab:arc_answer}
\end{table}

\begin{table}[h]
    \centering
     \resizebox{\columnwidth}{!}{\begin{tabular}{@{}>{\raggedright}p{4.6cm}cc@{}}
    \toprule
    \textbf{Model} & \textbf{Explainable}   & \textbf{Accuracy}  \\
    \midrule
       BERT$_{Large}$ & No  & 35.11 \\ 
     \midrule
     IR Solver~\cite{clark2016combining} & Yes & 20.26\\
     TupleILP~\cite{khot2017answering} & Yes  & 23.83   \\ 
     TableILP~\cite{khashabi2016question} & Yes   &  26.97  \\ 
    ExplanationLP\newline~\cite{thayaparan2021explainable} & Yes & 40.21 \\
     DGEM~\cite{clark2016combining} & Partial  &27.11  \\ 
     KG$^{2}$~\cite{zhang2018kg} & Partial   &  31.70 \\
     ET-RR~\cite{ni2018learning} & Partial & 36.61 \\
     Unsupervised AHE~\cite{yadav2019alignment} & Partial    &33.87 \\ 
     Supervised AHE~\cite{yadav2019alignment} & Partial  & 34.47 \\ 
     AutoRocc~\cite{yadav-etal-2019-quick} & Partial   & \textbf{41.24} \\ 
     \midrule
    \midrule
    \textit{Diff-}Explainer (ExplanationLP) & Yes  &\textbf{42.95}  \\ 
     \bottomrule
    \end{tabular}}
    \caption{\small ARC challenge scores compared with other Fully or Partially explainable approaches trained \textit{only} on the ARC dataset.}
    \label{tab:arc_baselines}
\end{table}

\subsection{Explanation Selection}

Table~\ref{tab:worldtree_explanation} shows the Precision@K scores for explanation retrieval for PathNet, ExplanationLP/TupleILP and \textit{Diff-}Explainer with ExplanationLP/TupleILP trained on answer and explanation selection. We choose Precision@K as the evaluation metric as the design of the approaches is not to construct full explanations but to take the top $k$=2 explanations and select the answer. 

As evident from the table, our approach significantly outperforms PathNet. We also improved the explanation selection performance over the non-differentiable solvers indicating the end-to-end fine-tuning also helps improve the selection of explanatory facts.

\begin{table}[t]
    \centering
    \small
  \resizebox{\linewidth}{!}{ 
  \begin{tabular}{p{3.5cm}cc}
    \toprule
    \textbf{Model} & Precision@1 & Precision@2 \\
    \midrule
     TupleILP& 40.44 & 31.21    \\ 
     ExplanationLP& 51.99 & 40.41     \\ 
     \midrule
     PathNet   & 19.79 & 13.73  \\
     \midrule
     \underline{\textbf{\textit{Diff-}Explainer}} \\
     TupleILP (Best)   & 40.64 & 32.23  \\
     ExplanationLP (Best)   & \textbf{56.77} & \textbf{41.91}  \\
     \bottomrule
    \end{tabular}
    }
    \caption{F1 score for explanation selection in WorldTree \textit{dev}-set .}
    \label{tab:worldtree_explanation}
\end{table}

\subsection{Answer Selection with Increasing Distractors}


\begin{table*}[t]
    \centering
    \small
    \resizebox{\textwidth}{!}{\begin{tabular}{p{16cm}}
    \toprule
    \textbf{Question (1):} Fanning can make a wood fire burn hotter because the fanning: 
    \textbf{Correct Answer:} adds more oxygen needed for burning. \\
    \underline{PathNet} \\
    \textbf{Answer}: provides the energy needed to keep the fire going.~\textbf{Explanations}: \textit{(i)} fanning a fire increases the oxygen near the fire, \textit{(ii)} placing a heavy blanket over a fire can be used to keep oxygen from reaching a fire \\
    \underline{ExplanationLP} \\
    \textbf{Answer}: increases the amount of wood there is to burn.~\textbf{Explanations}: \textit{(i)} more burning causes fire to be hotter, \textit{(ii)} wood burns \\
    \underline{\textit{Diff-}Explainer ExplanationLP} \\
    \textbf{Answer}: adds more oxygen needed for burning.~\textbf{Explanations}: \textit{(i)} more burning causes fire to be hotter, \textit{(ii)} fanning a fire increases the oxygen near the fire \\
    \midrule
    \textbf{Question (2):} Which type of graph would best display the changes in temperature over a 24 hour period? 
    \textbf{Correct Answer:} line graph. \\
    \underline{PathNet} \\
    \textbf{Answer}: circle/pie graph.~\textbf{Explanations}: \textit{(i)} a line graph is used for showing change ; data over time \\
    \underline{ExplanationLP} \\
    \textbf{Answer}: circle/pie graph.~\textbf{Explanations}: \textit{(i)} 1 day is equal to 24 hours, \textit{(ii)} a circle graph; pie graph can be used to display percents; ratios \\
    \underline{\textit{Diff-}Explainer ExplanationLP} \\
    \textbf{Answer}: line graph.~\textbf{Explanations}: \textit{(i)} a line graph is used for showing change; data over time, \textit{(ii)} 1 day is equal to 24 hours \\
     \midrule
    
    \textbf{Question (3):} Why has only one-half of the Moon ever been observed from Earth? 
    \textbf{Correct Answer:} The Moon rotates at the same rate that it revolves around Earth.. \\
    \underline{PathNet} \\
    \textbf{Answer}: The Moon has phases that coincide with its rate of rotation.~\textbf{Explanations}: \textit{(i)} the moon revolving around ; orbiting the Earth causes the phases of the moon,~\textit{(ii)} a new moon occurs 14 days after a full moon \\
    \underline{ExplanationLP} \\
    \textbf{Answer}: The Moon does not rotate on its axis.~\textbf{Explanations}: \textit{(i)} the moon rotates on its axis,~\textit{(ii)} the dark half of the moon is not visible \\
    \underline{\textit{Diff-}Explainer ExplanationLP} \\
    \textbf{Answer}: The Moon is not visible during the day.~\textbf{Explanations}: \textit{(i)} the dark half of the moon is not visible,~\textit{(ii)} a complete revolution; orbit of the moon around the Earth takes 1; one month \\
    \bottomrule
    \end{tabular}}
    \caption{Example of predicted answers and explanations (Only \textit{CENTRAL} explanations) obtained from our model with different levels of fine-tuning.}
    \label{tab:qa_examples_results}
\end{table*}

\begin{figure}[t]
    \centering
    \resizebox{\linewidth}{!}{\includegraphics{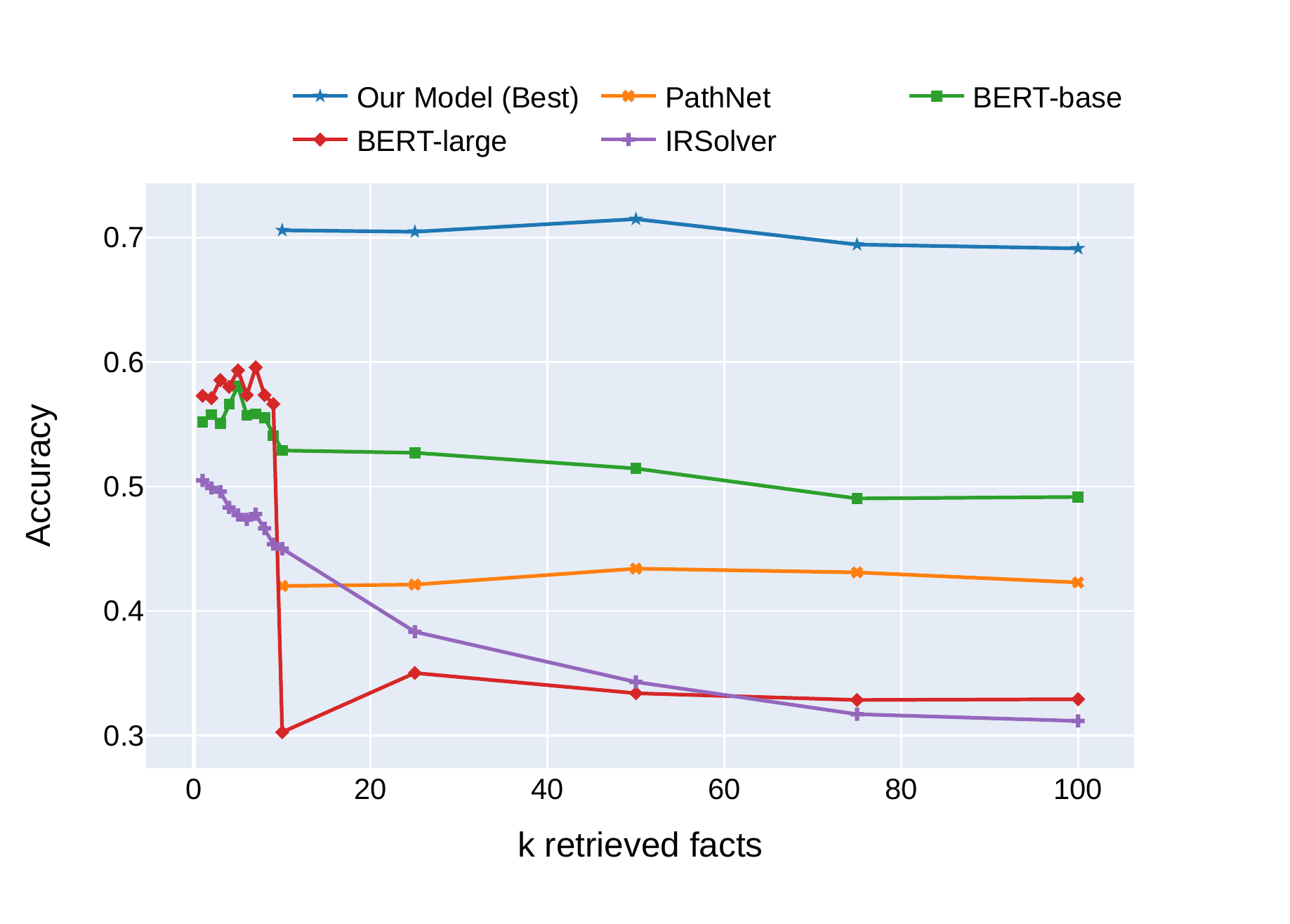}}
    \caption{Comparison of accuracy for different number of retrieved facts.}
    \label{fig:k_facts_plot}
\end{figure}

As noted by previous works~\cite{yadav-etal-2019-quick,yadav2020unsupervised}, the answer selection performance can decrease when increasing the number of used facts $k$ for Transformer. We evaluate how our approach stacks compared with transformer-based approaches in this aspect, presented in Figure~\ref{fig:k_facts_plot}.
As we can see, the IR Solver decreases in performance as we add more facts, while the scores for transformer-based models start deteriorating for $k>5$. Such results might seem counter-intuitive since it would be natural to expect a model's performance to increase as we add supporting facts. However, in practice, that does not apply as by adding more facts, there is an addition of distractors that such models may not filter out.

We can prominently see this for BERT$_{Large}$ with a sudden drop in performance for $k=10$, going from 56.61 to 30.26. Such drop is likely being caused by substantial overfitting; with the added noise, the model partially lost the ability for generalization. A softer version of this phenomenon is also observed for BERT$_{Base}$. 

In contrast,  our model's performance increases as we add more facts, reaching a stable point around $k=50$. Such performance stems from our combination of overlap and relevance scores along with the structural and semantic constraints. The obtained results highlight our model's robustness to distracting knowledge, allowing its use in data-rich scenarios, where one needs to use facts from extensive knowledge bases. PathNet is also exhibiting robustness across increasing distractors, but we consistently outperform it across all $k$ configurations.

On the other hand, for smaller values of $k$ our model is outperformed by transformer-based approaches, hinting that our model is more suitable for scenarios involving large knowledge bases such as the one presented in this work.





\subsection{Qualitative Analysis}



We selected some qualitative examples that showcase how end-to-end fine-tuning can improve the quality and inference and presented them in Table~\ref{tab:qa_examples_results}. We use the ExplanationLP for non-differentiable solver and \textit{Diff-}Explainer as they yield higher performance in answer and explanation selection.

For Question (1), \textit{Diff-}Explainer retrieves both explanations correctly and be able to answer correctly. Both PathNet and ExplanationLP has correctly retrieved at least one explanation but performed incorrect inference. We hypothesise that the other two approaches were distracted by the lexical overlaps in question/answer and facts, while our approach is robust towards distractor terms. In Question (2), our model was able only to retrieve one explanation correctly and was distracted by the lexical overlap to retrieve an irrelevant one. However, it still was able to answer correctly. In Question (3), all the approaches answered the question wrong, including our approach. Even though our approach was able to retrieve at least one correct explanation, it was not able to combine the information to answer and was distracted by lexical noise. These shortcomings indicate that more work can be done, and different constraints can be experimented with for combining facts.


%% file: sections/conclusion.tex
\section{Conclusion}
We presented a novel framework for encoding explicit and controllable assumptions as an end-to-end learning framework for question answering. We empirically demonstrated how incorporating these constraints in broader Transformer-based architectures can improve answer and explanation selection. The presented framework adopts constraints from TupleILP and ExplanationLP, but \textit{Diff-}Explainer can be extended to encode different constraints with varying degrees of complexity. 

This approach can also be extended to handle other forms of multi-hop QA, including open-domain, cloze style and answer generation. ILP has also been employed for relation extraction~\cite{roth-yih-2004-linear,choi-etal-2006-joint,chen-etal-2014-encoding}, semantic role labeling~\cite{punyakanok-etal-2004-semantic,koomen2005generalized}, sentiment analysis~\cite{choi2009adapting} and explanation regeneration~\cite{gupta-srinivasaraghavan-2020-explanation}. We can adapt and improve the constraints presented in this approach to build explainable approaches for the respective tasks. 

\textit{Diff-}Explainer is the first work investigating the intersection of explicit constraints and latent neural representations to the best of our knowledge. We hope this work will open the way for future lines of research on neuro-symbolic models, leading to more controllable, transparent and explainable NLP models.


\section*{Acknowledgements}

The work is partially funded by the EPSRC grant
EP/T026995/1 entitled ``EnnCore: End-to-End Conceptual Guarding of Neural Architectures” under \textit{Security for all in an AI enabled society}. The authors would like to thank the anonymous reviewers and editors of TACL for the constructive feedback. Additionally, we would like to thank the Computational Shared Facility of the University of Manchester for providing the infrastructure to run our experiments.

%% file: sections/Appendices.tex
\section{Appendix}

\subsection{Model Description}
\begin{figure*}[t]
    \centering
    \resizebox{0.9\textwidth}{!}{\includegraphics{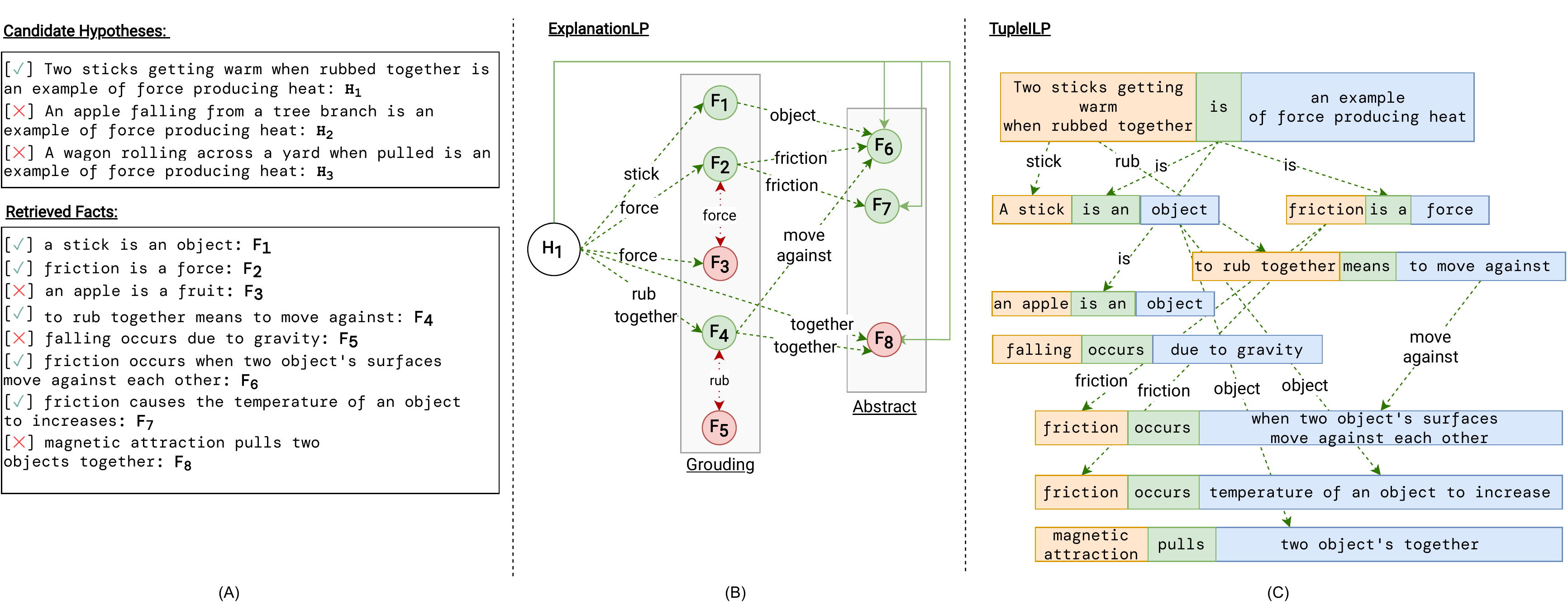}}
    \caption{ILP-Based Natural Language Inference}
    \label{fig:appendix}
\end{figure*}

This section presents a detailed explanation of TupleILP and ExplanationLP:

\paragraph{TupleILP} TupleILP uses Subject-Predicate-Object tuples for aligning and constructing the explanation graph. As shown in Figure~\ref{fig:appendix}C, the tuple graph is constructed and lexical overlaps are aligned to select the explanatory facts. The constraints are designed based on the position of text in the tuple. 

\paragraph{ExplanationLP} Given hypothesis $H_1$ from Figure~\ref{fig:appendix}A, the underlying concept the hypothesis attempts to test is the understanding of \textit{friction}. Different ILP approaches would attempt to build explanation graph differently. For example, ExplanationLP~\cite{thayaparan2021explainable} would classify core scientific facts ($F_6$-$F_8$) into \textit{abstract facts} and the linking facts  ($F_1$-$F_5$) that connects generic or abstract terms in the hypothesis into \textit{grounding fact}. The constraints are designed to emulate abstraction by starting to from the concrete statement to more abstract concepts via the grounding facts as shown in Figure~\ref{fig:appendix}B.

\subsection{Objective Function}

In this section, we explain how to design the objective function for TupleILP and ExplanationLP to adopt with \textit{Diff-}Explainer.

Given $n$ candidate hypotheses and $k$ candidate explanatory facts, $A$ represents an adjacency matrix of dimension $((n+k) \times (n+k))$ where the first $n$ columns and rows denote the candidate hypotheses, while the remaining rows and columns represent the candidate explanatory facts.  The adjacency matrix denotes the graph's lexical connections between hypotheses and facts.  Specifically, each entry in the matrix $A_{ij}$ contains the following values:

\begin{equation}
    \small
    A{ij}=
    \begin{cases}
      1, & i \leq n,j > n,~|trm(h_i)\cap trm(f_{j-n})| > 0 \\
      1, & j \leq n,i > n,~|trm(h_j)\cap trm(f_{i-n})| > 0 \\
      0, & \text{otherwise}
    \end{cases}
  \end{equation}

 \begin{table*}[t]
    \small
    \centering
    \resizebox{\textwidth}{!}{
    \begin{tabular}{p{4.5cm}>{\raggedright}p{6cm}p{7.5cm}}
    \toprule
   \textbf{Description} & \textbf{DPP Format} & \textbf{Parameters} \\ 
   \midrule
  \underline{\textbf{TupleILP}} \\ 
     Sub graph must have $\leq w_1$ active tuples & 
        \begin{equation}\sum_{i~\in~F} Y_{ii}  \leq w_1 + 1\end{equation} & - 
     \\
     Active hypothesis term must have $\leq w_2$ edges &

     \begin{equation}
         H_\theta[:,:,i] \odot Y \leq w_2 \quad \forall i~\in H^{t}
     \end{equation} &
     $H_\theta$ is  populated by hypothesis term matrix $H$ with dimension $((n+k) \times (n+k) \times l)$ and the values are given by:
\begin{equation}
    H_{ijk}=
    \begin{cases}
      1, & \parbox[t]{5.5cm}{$\forall ~k \in H^t, i \in H, j \in F,\\ t_k \in trm(h_i), t_k \in trm(f_{j})$} \\
      1, & \parbox[t]{5.5cm}{$\forall ~k \in H^t, i \in F, j \in H,\\ t_k \in trm(h_j), t_k \in trm(f_{i})$} \\
      0, & \text{otherwise}
    \end{cases}
\end{equation} \\
\midrule
     Active tuple must have active subject &
    \begin{equation}Y \odot T^S_{\theta} >= E \odot A_{\theta} \end{equation} & $A_{\theta}$ populated by adjacency matrix $A$, $T^S_{\theta}$ by subject tuple matrix $T^S$ with dimension $((n+k) \times (n+k))$ and the values are given by:
\begin{equation}
    T^S_{ij}=
    \begin{cases}
      1, & \parbox[t]{5.5cm}{$i \in H,~j \in F,\\  |trm(h_i) \cap trm(f^S_{j})| > 0$ }\\
      1, & \parbox[t]{5.5cm}{$i \in F,~j \in H,\\  |trm(h_j) \cap trm(f^S_{i})| > 0$ }\\
      0, & \text{otherwise}
    \end{cases}
\end{equation} 
     \\
     \midrule
     Active tuple must have $\geq w_3$ active fields & 
     
     \begin{equation}Y \odot T^S_{\theta} + Y \odot T^P_{\theta} + Y \odot T^O_{\theta} \geq w_3(Y \odot A_{\theta}) \end{equation}&

     $A_{\theta}$ populated by adjacency matrix $A$ and $T^S_{\theta}$, $T^P_{\theta}$, $T^O_{\theta}$  populated by subject, predicate and object matrices $T^S$, $T^P$, $T^O$  respectively. Predicate and object tuples are converted into $T^P,~T^O$ matrices similar to $T^S$   \\
     \midrule
     
     Active tuple must have an edge to some hypothesis term & Implemented during graph construction by only considering tuples that have lexical overlap with a hypothesis & - \\
     \midrule
     \midrule
    \underline{\textbf{ExplanationLP}} \\ 
    
    Limits the total number of abstract facts to $w_4$ & 
    
    \begin{equation}diag(Y) \cdot F^{AB}_{\theta}  \leq w_4\end{equation} 
    
    & $F^{AB}_{\theta}$ is populated by Abstract fact matrix $F^{AB}$, where:
    \begin{equation}
    F^{AB}_{ij}=
    \begin{cases}
      1, & \parbox[t]{5.5cm}{$i \in H,~j \in F^{A}$}\\
      0, & \text{otherwise}
    \end{cases}
\end{equation} 
    \\
    \bottomrule
    \end{tabular}}
    \caption{\small Adopting TupleILP and ExplanationLP  constraints in DPP format. For this work we set the hyperparameters $w_1$=2, $w_2$=2, $w_3$=1 and $w_4$=2}
    \label{tab:tupleilp_constraints}
\end{table*}

Given the relevance scoring functions, we construct edge weights matrix ($W$) via a weighted function for each approach as follows:

\paragraph{TupleILP} The weight function for \textit{Diff-}Explainer with TupleILP constraints is: 
\begin{align}
    \small
    W_{ij}= (\theta_{s_r} S_{ij} + \theta_{l_r} L_{ij}) \times A_{ij}~\forall i,j \in V 
\end{align}

\paragraph{ExplanationLP} Give Abstract KB ($F_A)$ and Grounding KB ($F_G)$, the weight function for \textit{Diff-}Explainer with Explanation LP is as follows:

     \begin{equation}
\resizebox{\columnwidth}{!}{%
$W_{ij} =
\begin{cases}
-\theta_{gg}L_{ij} & v_j,v_k\in F_{G}\\
-\theta_{aa} L_{ij}  & v_j, v_k \in F_{A}\\
\theta_{ga}L_{ij}  & v_j \in F_{G}, v_k \in F_{A}\\
\theta_{qgl}L_{ij} + \theta_{qgs} S_{ij}  & v_j \in F_{G}, v_k = {h_i}\\
\theta_{qal}L_{ij} + \theta_{qal}S_{ij} & v_j \in F_{A}, v_k = {h_i} \\
\end{cases}$
}
\end{equation}

\subsection{Constraints with Disciplined Parameterized Programming (DPP)}
\label{sec:dpp}



In order to adopt differentiable convex optimization layers, the constraints should be defined following the Disciplined Parameterized Programming (DPP) formalism~\cite{agrawal2019differentiable}, providing a set of conventions when constructing convex optimization problems. DPP consists of functions (or \textit{atoms}) with a known curvature (affine, convex or concave) and per-argument monotonicities. In addition to these, DPP also consists of \emph{Parameters} which are symbolic constants with an unknown numerical value assigned during the solver run.


\paragraph{TupleILP}  We extract SPO tuples $f^t_i = \{f^S_i,~f^P_i,~f^O_i\}$ for each fact $f_i$  using an Open Information Extraction model~\cite{stanovsky-etal-2018-supervised}. From the hypothesis $h_i$ we extract the set of unique terms $h^{ht}_i = \{t^{h_i}_1,~t^{h_i}_2,~t^{h_i}_3,~\dots,~t^{h_i}_l\}$ excluding stopwords. 

In addition to the aforementioned constraints and semidefinite constraints specified in Equation~\ref{eq:sdp_constraint}, we adopt part of the constraints from TupleILP~\cite{khot2017answering}. In order to implement TupleILP constraints, we extract SPO tuples $f^t_i = \{f^S_i,~f^P_i,~f^O_i\}$ for each fact $f_i$  using an Open Information Extraction model~\cite{stanovsky-etal-2018-supervised}. From the hypotheses $H$ we also extract the set of unique terms $H^{t} = \{t_1,~t_2,~t_3,~\dots,~t_l\}$ excluding stopwords. The constraints are described in Table~\ref{tab:tupleilp_constraints}. 

\paragraph{ExplanationLP} ExplanationLP constraints are described in Table~\ref{tab:tupleilp_constraints}.